\documentclass{article}

\PassOptionsToPackage{numbers, compress}{natbib}

\usepackage[final]{nips_2016}
\usepackage[utf8]{inputenc} 
\usepackage[T1]{fontenc}    
\usepackage[english]{babel}
\usepackage{url}            
\usepackage{booktabs}       
\usepackage{amsfonts}       
\usepackage{amsmath}        
\usepackage{nicefrac}       
\usepackage{microtype}      
\usepackage[table]{xcolor}			
\usepackage{graphicx}
\usepackage{caption}
\usepackage{subcaption}
\usepackage{wrapfig}
\usepackage{bm}
\usepackage[]{algorithm2e} 

\usepackage[switch,mathlines]{lineno}
\usepackage{linenofix} 

\usepackage[colorlinks]{hyperref}
\definecolor{mydarkblue}{rgb}{0,0.08,0.45}
\hypersetup{ %
    pdftitle={},
    pdfauthor={},
    pdfsubject={},
    pdfkeywords={},
    pdfborder=0 0 0,
    pdfpagemode=UseNone,
    colorlinks=true,
    linkcolor=mydarkblue,
    citecolor=mydarkblue,
    filecolor=mydarkblue,
    urlcolor=mydarkblue,
    pdfview=FitH}

\addto\extrasenglish{
}

\usepackage{xargs}
\definecolor{OliveGreen}{rgb}{0.33, 0.42, 0.18}
\definecolor{Plum}{rgb}{0.56, 0.27, 0.52}
\usepackage[colorinlistoftodos,textsize=small,disable]{todonotes}  
\newcommandx{\unsure}[2][1=]{\todo[linecolor=red,backgroundcolor=red!25,bordercolor=red,#1]{#2}}
\newcommandx{\change}[2][1=]{\todo[linecolor=blue,backgroundcolor=blue!25,bordercolor=blue,#1]{#2}}
\newcommandx{\info}[2][1=]{\todo[linecolor=OliveGreen,backgroundcolor=OliveGreen!25,bordercolor=OliveGreen,#1]{#2}}
\newcommandx{\improvement}[2][1=]{\todo[linecolor=Plum,backgroundcolor=Plum!25,bordercolor=Plum,#1]{#2}}
\title{Tagger: Deep Unsupervised Perceptual Grouping}

\author{
  Klaus Greff\textsuperscript{*}, Antti Rasmus, Mathias Berglund, Tele Hotloo Hao,\\\textbf{Jürgen Schmidhuber\textsuperscript{*}, Harri Valpola}\\
  The Curious AI Company \texttt{\{antti,mathias,hotloo,harri\}@cai.fi}\\
  \textsuperscript{*}IDSIA \texttt{\{klaus,juergen\}@idsia.ch}
}

\begin{document}
\maketitle
\begin{abstract}
We present a framework for efficient perceptual inference that explicitly reasons about the segmentation of its inputs and features. 
Rather than being trained for any specific segmentation, our framework learns the grouping process in an unsupervised manner or alongside any supervised task.
We enable a neural network to group the representations of different objects in an iterative manner through a differentiable mechanism. 
We achieve very fast convergence by allowing the system to amortize the joint iterative inference of the groupings and their representations. 
In contrast to many other recently proposed methods for addressing multi-object scenes, our system does not assume the inputs to be images and can therefore directly handle other modalities.
We evaluate our method on multi-digit classification of very cluttered images that require texture segmentation.
Remarkably our method achieves improved classification performance over convolutional networks despite being fully connected, by making use of the grouping mechanism.
Furthermore, we observe that our system greatly improves upon the semi-supervised result of a baseline Ladder network on our dataset.
These results are evidence that grouping is a powerful tool that can help to improve sample efficiency.
\end{abstract}


\section{Introduction}
\begin{wrapfigure}[20]{r}{0.28\textwidth}
\centering
\includegraphics[width=0.26\textwidth]{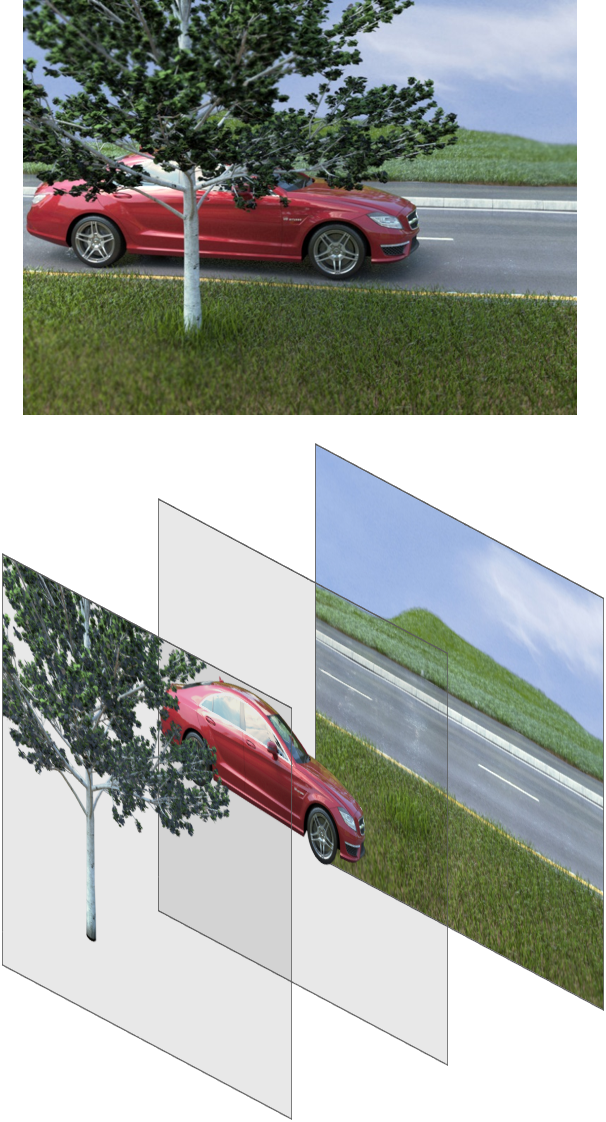}
\caption{An example of perceptual grouping for vision.}
\label{fig:seg_example}
\end{wrapfigure}
Humans naturally perceive the world as being structured into different objects, their properties and relation to each other. 
This phenomenon which we refer to as perceptual grouping is also known as amodal perception in psychology.
It occurs effortlessly and includes a segmentation of the visual input, such as that shown in in \autoref{fig:seg_example}. 
This grouping also applies analogously to other modalities, for example in solving the cocktail party problem (audio) or when separating the sensation of a grasped object from the sensation of fingers touching each other (tactile).
Even more abstract features such as object class, color, position, and velocity are naturally grouped together with the inputs to form coherent objects.
This rich structure is crucial for many real-world tasks such manipulating objects or driving a car, where awareness of different objects and their features is required.

In this paper, we introduce a framework for learning efficient iterative inference of such perceptual grouping which we call \emph{iTerative Amortized Grouping} (TAG). 
This framework entails a mechanism for iteratively splitting the inputs and internal representations into several different groups. 
We make no assumptions about the structure of this segmentation and rather train the model end-to-end to discover which are the relevant features and how to perform the splitting.

By using an auxiliary denoising task we focus directly on amortizing the posterior inference of the object features and their grouping.
Because our framework does not make any assumptions about the structure of the data, it is completely domain agnostic and applicable to any type of data.
The TAG framework works completely unsupervised, but can also be combined with supervised learning for classification or segmentation.

Another class of recently proposed mechanisms for addressing complex structured inputs is attention~\cite{Schmidhuber1991,Bahdanau2014,Eslami2016}. 
These methods simplify the problem of perception by learning to restrict processing to a part of the input. 
In contrast, TAG simply structures the input without directing the focus or discarding irrelevant information. 
These two systems are not mutually exclusive and could complement each other: the group structure can help in deciding what exactly to focus on, which in turn may help simplify the task at hand.

We apply our framework to two artificial datasets: a simple binary one with multiple shapes and one with overlapping textured MNIST digits on a textured background. 
We find that our method learns intuitively appealing groupings that support denoising and classification.
Our results for the 2-digit classification are significantly better than a strong ConvNet baseline despite the use of a fully connected network.
The improvements for semi-supervised learning with 1,000 labels are even greater, suggesting that grouping can help learning by increasing the sample efficiency.


\section{Iterative Amortized Grouping (TAG)}


\paragraph{Grouping.}

Our goal is to enable neural networks to split inputs and internal representations into coherent groups that can be processed separately.
We hypothesize that processing the whole input in one clump is often difficult due to unwanted interference.
However, if we allow the network to separately process groups, it can make use of invariant distributed features without the risk of ambiguities.
We thus define a group to be a collection of inputs and internal representations that are processed together (largely) independently of the other groups.

The ``correct'' grouping is often dynamic, ambiguous and task dependent. 
For example, when driving along a road, it is useful to group all the buildings together.
To find a specific house, however, it is important to separate the buildings, and to enter one, they need to be subdivided even further. 
Rather than treating segmentation as a separate task, we provided a mechanism for grouping as a tool for our system.
We make no assumptions about the correspondence between objects and groups. 
If the network can process several objects in one group without unwanted interference, then the network is free to do so.


Processing of the input is split into $K$ different groups, but it is left up to the network to learn how to best use this ability in a given problem, such as classification. 
To make the task of instance segmentation easy, we keep the groups symmetric in the sense that each group is processed by the same underlying model.
We introduce latent binary variables $g_{k,j}$ to encode if input element $x_j$ is assigned to group $k$.%
\footnote{More formally, we introduce discrete random variables $G_j$ for each input element indexed by $j$. 
As a shorthand for $p(G_j = k)$ we write $p(g_{k, j})$ and denote the discrete-valued vector of elements $g_{k,j}$ by $\bm{g}_k$.}

\paragraph{Amortized Iterative Inference.}

We want our model to reason not only about the group assignments but also about the representation of each group. 
This amounts to inference over two sets of variables: the latent group assignments and the individual group representations; 
A formulation very similar to mixture models for which exact inference is typically intractable.
For these models it is a common approach to approximate the inference in an iterative manner by alternating between (re-)estimation of these two sets (e.g., EM-like methods~\cite{Dempster1977}).
The intuition is that given the grouping, inferring the object features becomes easy, and vice versa.
We employ a similar strategy by allowing our network to iteratively refine its estimates of the group assignments as well as the object representations.
If the model can improve the estimates in each step, then it will converge to a final solution.


Rather than deriving and then running an inference algorithm, we train a parametric mapping to arrive at the end result of inference as efficiently as possible~\cite{Gregor2010}. 
This is known as \emph{amortized} inference~\cite{Srikumar2012}, and it is used, for instance, in variational autoencoders where the encoder learns to amortize the posterior inference required by the generative model represented by the decoder.
Here we instead apply the framework of denoising autoencoders \cite{Gallinari1987,LeCun1987,Vincent2008} which are trained to reconstruct original inputs $\bm{x}$ from corrupted versions $\bm{\tilde{x}}$. 
This encourages the network to implement useful amortized posterior inference without ever having to specify or even know the underlying generative model whose inference is implicitly learned. 

This situation is analogous to normal supervised deep learning, which can also be viewed as amortized inference~\cite{Bengio2014}. 
Rather than specifying all the hidden variables that are related to the inputs and labels and then deriving and running an inference algorithm, a supervised deep model is trained to arrive at an approximation $q(\mathrm{class} \mid \mathrm{input})$ of the true posterior $p(\mathrm{class} \mid \mathrm{input})$ without the user specifying or typically even knowing the underlying generative model. 
This works as long as the network is provided with the input information and mechanisms required for an an efficient approximation of posterior inference.


\subsection{Definition of the TAG mechanism}
\label{sec:synthesis}

\begin{figure}
\centering
\includegraphics[width=1.0\textwidth]{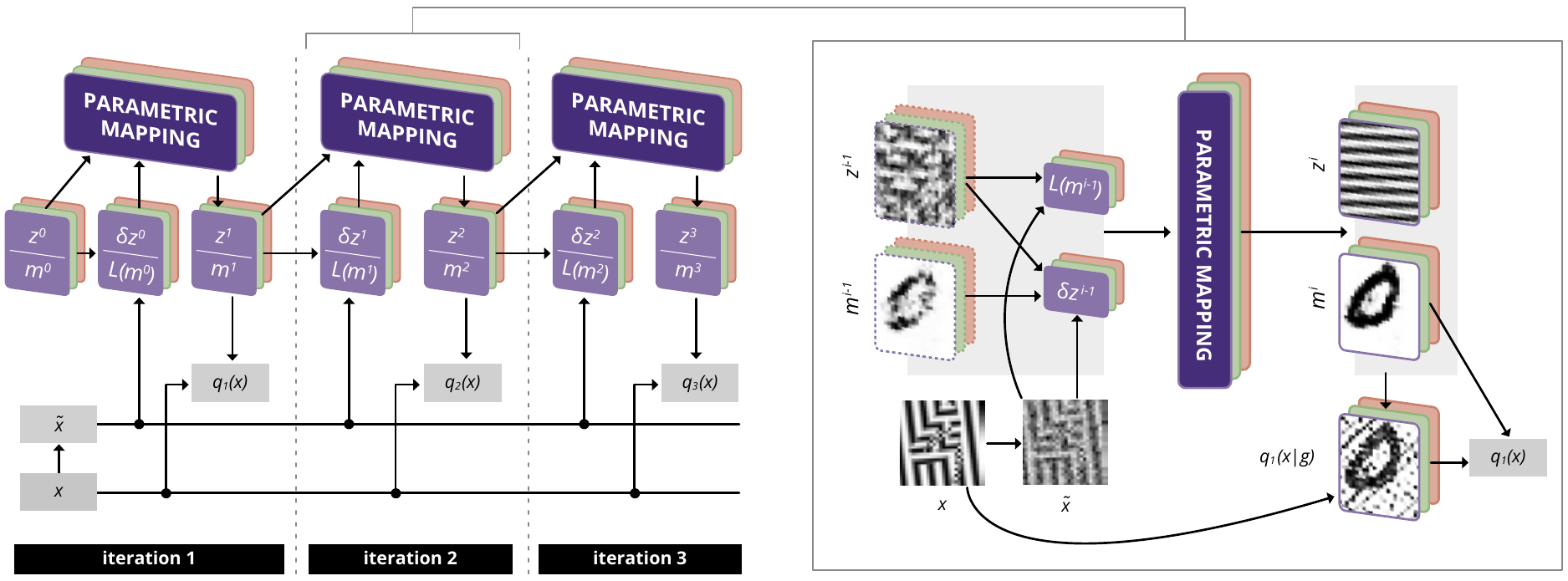}
\caption{Illustration of the TAG framework used for training. Left: The system
learns by denoising its input over iterations using several groups to distribute the
representation.
Each group, represented by several panels of the same color, maintains its own estimate of reconstructions $\bm{z}^i$ of the input, and corresponding masks $\bm{m}^i$, which encode the parts of the input that this group is responsible for representing. 
These estimates are updated over iterations by the same network, that is, each group and iteration share the weights of the network and only the inputs to the network differ.
In the case of images, $\bm{z}$ contains pixel-values.
Right: In each iteration $\bm{z}^{i-1}$ and $\bm{m}^{i-1}$ from the previous iteration, 
are used to compute a likelihood term $L(\bm{m}^{i-1})$ and modeling error $\delta \bm{z}^{i-1}$. 
These four quantities are fed to the parametric mapping to produce $\bm{z}^i$ and $\bm{m}^i$ for the next iteration. 
During learning, all inputs to the network are derived from the corrupted input as shown here. 
The unsupervised task for the network is to learn to denoise, i.e. output an estimate $q(\bm{x})$ of the original clean input.
See \autoref{sec:synthesis} for more details.}
\label{fig:tagger}
\end{figure}

A high-level illustration of the TAG framework is presented in \autoref{fig:tagger}: 
We train a network with a learnable grouping mechanism to iteratively denoise corrupted inputs $\bm{\tilde{x}}$.
The output at each iteration is an approximation $q_i(\bm{x})$ of the true probability $p(\bm{x} \mid \bm{\tilde{x}})$, which is refined over iterations indexed by $i$. 
As the cost function for training the network, we use the negative log likelihood 
\begin{align} \label{eq:cost}
C(\bm{x}) = -\sum_i \log q_i(\bm{x}),
\end{align}
where the summation is over iterations $i$. From here on we mostly omit $i$ from the equations for readability.
Since this cost function does not require any class labels or intended grouping information, training can be completely unsupervised, though additional terms for supervised tasks can be added too.

\paragraph{Group representation.}

Internally, the network maintains $K$ versions of its representations indexed by $k$. 
This can also be thought of as running $K$ separate copies of the same network, where each network only sees a subset of the inputs and outputs $\bm{z}_k = q(\bm{x} | \bm{g}_k)$ (the expected value of the input for that group), and $m_k = q(\bm{g}_k)$ (the group assignment probabilities). 
Each $\bm{z}_k$ and $\bm{m}_k$ has the same dimensionality as the input, and they are updated over iterations.
Each group $k$ makes its own prediction about the original input based on $\bm{z}_k$. 
In the binary case we use $q(x_j \mid g_{k,j}) = \operatorname{sigmoid}(z_{k,j})$, and in the continuous case we take $z_{k,j}$ to represent the mean of a Gaussian distribution with variance $v$.
We assumed the variance of the Gaussian distribution to be constant over iterations and groups but learned it from the data. It would be easy to add a more accurate estimate of the variance.

The final prediction of the network is defined as: 
\begin{align}
q(x_j) = \sum_k q(g_{k,j}) q(x_j \mid g_{k,j}) = \sum_k m_{k,j} q(x_j \mid g_{k,j}).
\end{align}
The group assignment probabilities $q(g_{k,j}) = \bm{m}_k$ are forced to be non-negative and sum up to one over $k$: 
\begin{align} \label{eq:m_conditions}
m_{k,j} \geq 0, \hspace{1cm} \sum_k m_{k,j} = 1.
\end{align}

\paragraph{Inputs.}

In contrast to a normal denoising autoencoder that receives the corrupted $\bm{\tilde{x}}$, we feed in estimates $\bm{m}^i_k$ and $\bm{z}^i_k$ from the previous iteration and two additional quantities: $L(\bm{m}^i_k)$ and $\bm{\delta z}^i_k$.
They are functions of the estimates and the corrupted $\bm{\tilde{x}}$ and carry information about how the estimates could be improved.
A parametric mapping (here a neural network) then produces the new estimates $\bm{m}^{i+1}_k$ and $\bm{z}^{i+1}_k$.
The initial values for $\bm{m}_k^0$ are randomized, and $\bm{z}_k^0$ is set to the data mean for all $k$.

Because $\bm{z}_k$ are continuous variables, their likelihood is a function over all possible values of $\bm{z}_k$, and not all of this information can be easily represented. 
Typically, the relevant information is found close to the current estimate $\bm{z}_k$; therefore we use $\delta \bm{z}_k$, which is proportional to the gradient of the negative log likelihood.
Essentially, it represents the remaining modeling error:
\begin{equation} \label{eq:delta_z}
\delta z_{k,j} = m_{k,j} (\tilde{x}_j - z_{k, j}) \propto \frac{\partial}{\partial z_{k,j}}
  \left[ -\log (\sum_h q(\tilde{x}_j \mid z_{h, j}, g_{h,j})) \right] \, .
\end{equation}
The derivation of the analogous term in the binary case is presented in Appendix~\ref{sec:deltaz}.

Since we are using denoising as a training objective, the network can only be allowed to take inputs through the corrupted $\bm{\tilde{x}}$ during learning.
Therefore, we need to look at the likelihood $p(\bm{\tilde{x}} \mid \bm{z}_:, \bm{g}_:)$ of the corrupted input when trying to determine how the estimates could be improved.
Since $g_{k,j}$ are discrete variables unlike $z_{k,j}$, we treat them slightly differently: 
For $g_{k,j}$ it is feasible to express the complete likelihood table assuming other values constant.
We denote this function by
\begin{equation}
L(m_{k, j}) \propto q(\tilde{x}_j \mid z_{:, j}, g_{:, j}) \, .
\end{equation}
Note that we normalize $L(m_{k, j})$ over $k$ such that it sums up to one for each value of $j$. 
This amounts to providing each group information about how likely each input element belongs to them rather than some other group. 
In other words, this is equivalent to likelihood ratio rather than the raw likelihood.
Intuitively, the term $L(\bm{m}_k)$ describes how well each group reconstructs the individual input elements relative to the other groups.

\paragraph{Parametric mapping.}
The final component needed in the TAG framework is the parametric model, which does all the heavy lifting of inference.
This model has a dual task: first, to denoise the estimate $\bm{z}_k$ of what each group says about the input, and second, to update the group assignment probabilities $\bm{m}_k$ of each input element. 
The information about the remaining modeling error is based on the corrupted input $\bm{\tilde{x}}$; thus, the parametric network has to denoise this and in effect implement posterior inference for the estimated quantities. 
The mapping function is the same for each group $k$ and for each iteration.
In other words, we share weights and in effect have only a single function approximator that we reuse.

The denoising task encourages the network to iteratively group its inputs into coherent groups that can be modeled efficiently.
The trained network can be useful for a real-world denoising application, but typically, the idea is to encourage the network to learn interesting internal representations.
Therefore, it is not $q(\bm{x})$ but rather $\bm{m}_k$, $\bm{z}_k$ and the internal representations of the parametric mapping that we are typically concerned with.

\newpage
\paragraph{Summary.}
By using the negative log likelihood $C(\bm{x}) = - \sum_i \log q_i(\bm{x})$ as a cost function, we train our system to compute an approximation $q_i(\bm{x})$ of the true denoising posterior $p(\bm{x}|\bm{\tilde{x}})$ at each iteration $i$.
An overview of the whole system is given in \autoref{fig:tagger}.
For each input element $x_j$ we introduce $K$ latent binary variables $g_{k,j}$ that take a value of $1$ if this element is generated by group $k$.
This way inference is split into $K$ groups, and we can write the approximate posterior in vector notation as follows:
\begin{equation}
  q_i(\bm{x}) = \sum_k q_i(\bm{x}|\bm{g}_k) q_i(\bm{g}_k) = \sum_k \mathcal{N}(\bm{x}; \bm{z}_k^i, v\bm{I}) \bm{m}_k^i\;,
\label{eq:model}
\end{equation}
where we model the group reconstruction $q_i(\bm{x}|\bm{g_k})$ as a Gaussian with mean $\bm{z}_k^i$ and variance $v$, and the group assignment posterior $q_i(\bm{g_k})$ as a categorical distribution $\bm{m}_k$.

The trainable part of the TAG framework is given by a parametric mapping that operates independently on each group $k$ and is used to compute both $\bm{z}_k^i$ and $\bm{m}_k^i$ (which is afterwards normalized using an elementwise softmax over the groups).
This parametric mapping is usually implemented by a neural network and the whole system is trained end-to-end using standard backpropagation through time. 

The input to the network for the next iteration consists of the vectors $\bm{z}_k^{i}$ and $\bm{m}_k^{i}$ along with two additional quantities: 
The remaining modelling error $\bm{\delta z}_k^i$ and the group assignment likelihood ratio $L(\bm{m}_k^i)$ which carry information about how the estimates can be improved: 
\[
  \bm{\delta z}_k^i \propto \frac{\partial C(\bm{\tilde{x}})}{\partial \bm{z}_k^i}
  \qquad\text{and}\qquad
  L(\bm{m}_k^i) \propto \frac{q_i(\bm{\tilde{x}} | \bm{g}_k)}{\sum_h q_i(\bm{\tilde{x}} | \bm{g}_h)}
\]
Note that they are derived from the corrupted input $\bm{\tilde{x}}$, to make sure we don't leak information about the clean input $\bm{x}$ into the system.

\subsection{The Tagger: Combining TAG and Ladder Network}
\label{seq:ladder}
\begin{figure}
\centering
\begin{minipage}{0.55\textwidth}
\centering
\begin{algorithm}[H]
  \KwData{$\bm{x}, K, T, \sigma, v, W_h, W_u, \Theta$}
  \KwResult{$\bm{z}^T, \bm{m}^T,  C$}
  \Begin(Initialization:){
    $\bm{\tilde{x}} \leftarrow \bm{x} + \mathcal{N}(\bm{0}, \sigma^2\bm{I}) $\;
    $\bm{m}^0 \leftarrow \text{softmax}(\mathcal{N}(\bm{0}, \bm{I}))$\;
    $\bm{z}^0 \leftarrow E[\bm{x}]$\;
  }
  \For{$i = 0 \dots T-1$}{
    \For{$k = 1 \dots K$}{
      $\bm{\tilde{z}}_k \leftarrow \mathcal{N}(\bm{\tilde{x}}; \bm{z}^i_k, (v + \sigma^2) \bm{I})$\;
      $\bm{\delta z^i_k} \leftarrow (\bm{\tilde{x}} - \bm{z}_k^i) \bm{m}^i_k \bm{\tilde{z}}_k $\;
      $L(\bm{m}^i_k) \leftarrow \frac{\bm{\tilde{z}}_k}{\sum_h \bm{\tilde{z}}_h}$ \;
      $\bm{h}_k^i \leftarrow f(W_h \left[ \bm{z}^i_k, \bm{m}^i_k, \bm{\delta z}^i_k, L(\bm{m}^i_k) \right])$\;
      $\bm{u}^{i}_k \leftarrow  \text{Ladder}(\bm{h}_k^i, \Theta)$\;
      $[ \bm{z}^{i+1}_k, \bm{m}^{i+1}_k ] \leftarrow W_u \bm{u}^{i}_k$\; 
    }
    $\bm{m}^{i+1} \leftarrow \operatorname{softmax} (\bm{m}^{i+1})$\;
    $q_{i+1}(\bm{x}) \leftarrow \sum_{k=1}^K \mathcal{N}(\bm{x}; \bm{z}^{i+1}_k, v\bm{I}) \bm{m}^{i+1}$\;
  }
 $C \leftarrow - \sum_{i=1}^T \log q_i(\bm{x}) $\;
 \vspace{0.5em}
 \caption{Pseudocode for running Tagger on a single real-valued example $\bm{x}$. For details and a binary-input version please refer to supplementary material.}
 \label{alg:tagger}
\end{algorithm}
\end{minipage}\hfill
\begin{minipage}{0.42\textwidth}
    \centering
    \includegraphics[width=\textwidth]{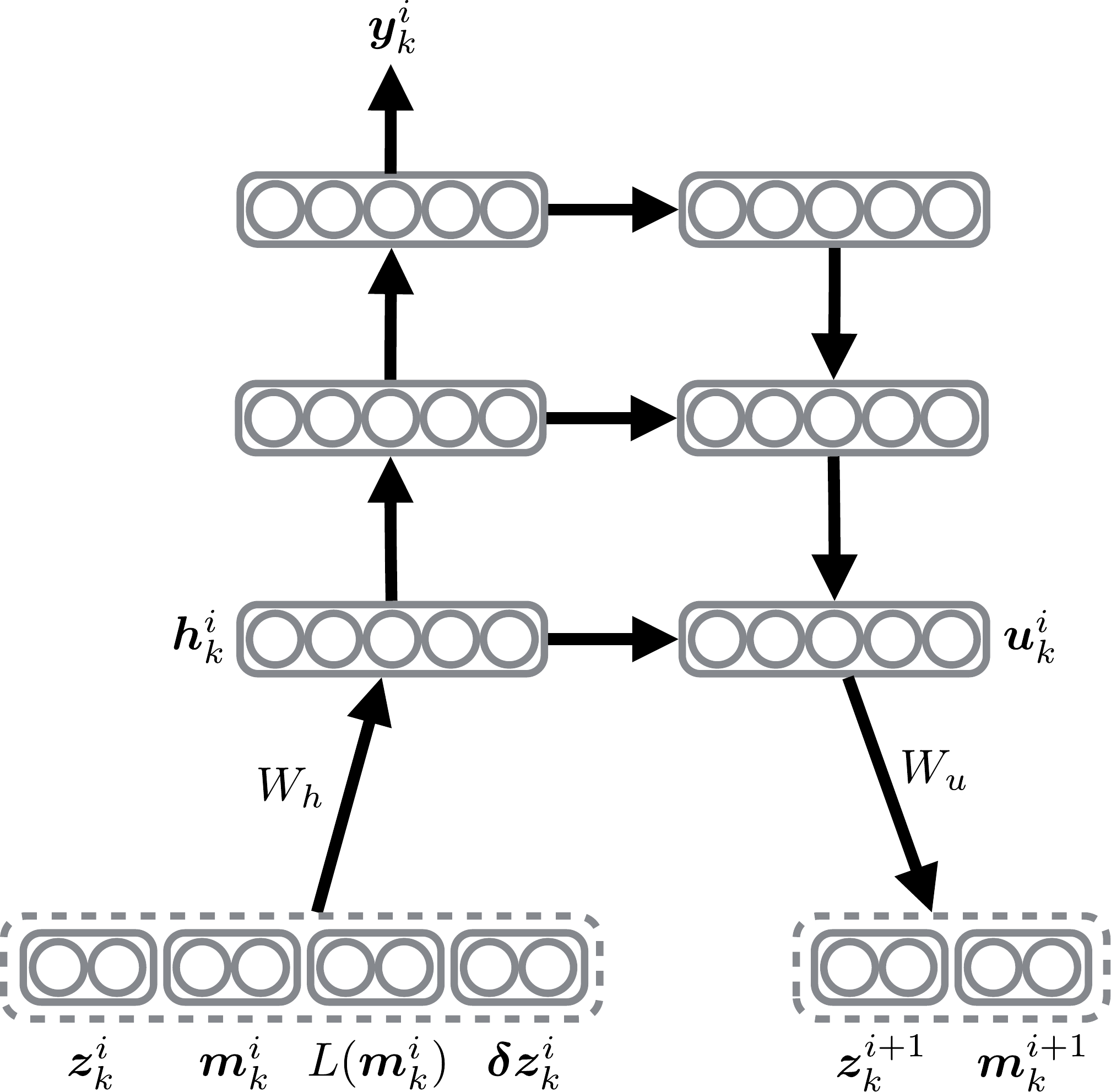}
    \caption{An example of how Tagger would use a 3-layer-deep Ladder Network as
    its parametric mapping to perform its iteration $i+1$. Note the optional 
    class prediction output   
    $\bm{y}^i_g$ for classification tasks. See supplementary material for details.}
    \label{fig:ladder}
\end{minipage}
\vspace{-1em}
\end{figure}

We chose the Ladder network \cite{Rasmus2015} as the parametric mapping because its structure reflects the computations required for posterior inference in hierarchical latent variable models. 
This means that the network should be well equipped to handle the hierarchical structure one might expect to find in many domains. 
We call this Ladder network wrapped in the TAG framework \emph{Tagger}.
This is illustrated in \autoref{fig:ladder} and the corresponding pseudocode can be found in \autoref{alg:tagger}.

We mostly used the specifications of the Ladder network as described by \citet{Rasmus2015}, but there are some minor modifications we made to fit it to the TAG framework. 
We found that the model becomes more stable during iterations when we added a sigmoid function to the gating variable $v$~\citep[Equation 2]{Rasmus2015} used in all the decoder layers with continuous outputs.
None of the noise sources or denoising costs were in use (i.e., $\lambda_l=0$ for all $l$ in Eq.~3 of Ref.~\cite{Rasmus2015}), but Ladder's classification cost ($C_c$ in Ref.~\cite{Rasmus2015}) was added to the Tagger's cost (\autoref{eq:cost}) for the semi-supervised tasks.

All four inputs ($\bm{z}^i_k$, $\bm{m}^i_k$, $\delta \bm{z}^{i}_k$, and $L(\bm{m}^{i}_k)$) were concatenated and projected to a hidden representation that served as the input layer of the Ladder Network. 
Subsequently, the values for the next iteration were simply read from the reconstruction ($\hat{\bm{x}}$ in Ref.~\cite{Rasmus2015}) and projected linearly into $\bm{z}^{i+1}_k$ and via softmax to $\bm{m}^{i+1}_k$ to enforce the conditions in \autoref{eq:m_conditions}. 
For the binary case, we used a logistic sigmoid activation for $\bm{z}^{i+1}_k$.


\section{Experiments and results}

We explore the properties and evaluate the performance of Tagger both in fully unsupervised settings and in semi-supervised tasks in two datasets\footnote{The datasets and a Theano~\cite{TheTheanoDevelopmentTeam2016} reference implementation of Tagger are available at \url{http://github.com/CuriousAI/tagger}}.
Although both datasets consist of images and grouping is intuitively similar to image segmentation, there is no prior in the Tagger model for images: our results (unlike the ConvNet baseline) generalize even if we permute all the pixels .

\paragraph{Shapes.}

We use the simple Shapes dataset~\cite{Reichert2011} to examine the basic properties of our system. 
It consists of 60,000 (train) + 10,000 (test) binary images of size 20x20.
Each image contains three randomly chosen shapes ($\bigtriangleup$, $\bigtriangledown$, or $\square$) composed together at random positions with possible overlap.

\paragraph{Textured MNIST.}

We generated a two-object supervised dataset (TextureMNIST2) by sequentially stacking two textured 28x28 MNIST-digits, shifted two pixels left and up, and right and down, respectively, on top of a background texture.
The textures for the digits and background are different randomly shifted samples from a bank of 20 sinusoidal textures with different frequencies and orientations.
Some examples from this dataset are presented in the column of \autoref{fig:mnist_qualitative}.
We use a 50k training set, 10k validation set, and 10k test set to report the results.
The dataset is assumed to be difficult due to the heavy overlap of the objects in addition to the clutter due to the textures. 
We also use a textured single-digit version (TextureMNIST1) without a shift to isolate the impact of texturing from multiple objects. 

\subsection{Training and evaluation}

We train Tagger in an unsupervised manner by only showing the network the raw input example $\bm{x}$, not ground truth masks or any class labels, using 4 groups and 3 iterations.
We average the cost over iterations and use ADAM~\cite{Kingma2015} for optimization.
On the Shapes dataset we trained for 100 epochs with a bit-flip probability of 0.2, and on the TextureMNIST dataset for 200 epochs with a corruption-noise standard deviation of 0.2.
The models reported in this paper took approximately 3 and 11 hours in wall clock time on a single Nvidia Titan X GPU for Shapes and TextureMNIST2 datasets respectively.

To understand how model size, length of the iterative inference, and the number of groups affect the modeling performance, we evaluate the trained models using two metrics:
First, the denoising cost on the validation set, and second we evaluate the segmentation into objects using the \emph{adjusted mutual information (AMI) score} \cite{Vinh2010} and ignore the background and overlap regions in the Shapes dataset (consistent with~\citet{Greff2015}). 
Evaluations of the AMI score and classification results in semi-supervised tasks were performed using uncorrupted input.
The system has no restrictions regarding the number of groups and iterations used for training and evaluation.
The results improved in terms of both denoising cost and AMI score when iterating further, so we used 5 iterations for testing.
Even if the system was trained with 4 groups and 3 shapes per training example, we could test 
the evaluation with, for example, 2 groups and 3 shapes, or 4 groups and 4 shapes.

\subsection{Unsupervised Perceptual Grouping}
\begin{table}
	\centering
\begin{subtable}[t]{0.5\linewidth}
\centering
  \begin{tabular}{ r l l l l l l}
          & Iter 1 & Iter 2 & Iter 3 & Iter 4 & Iter 5\\
  \toprule
  Denoising cost & 0.094  & 0.068 &  {0.063} & {0.063} & {0.063}\\ 
  AMI & 0.58 & 0.73 & 0.77 & {0.79} & {0.79}\\  
  \midrule

  Denoising cost* & 0.100  & 0.069 &  0.057 & \textbf{0.054} & \textbf{0.054}\\ 
  AMI* &  0.70 & 0.90 & 0.95 & 0.96 & \textbf{0.97}\\
  \end{tabular}
  \caption{Convergence of Tagger over iterative inference}
  \label{tab:results_shapes_a}
\end{subtable}
\hspace{7em}
\begin{subtable}[t]{0.3\linewidth}
  	\centering
    \begin{tabular}{ r c}
          & AMI\\
          \toprule
          RC \cite{Greff2015} & 0.61 $\pm$ 0.005\\   
          Tagger & 0.79 $\pm$ 0.034\\  
          Tagger* & \textbf{0.97 $\pm$ 0.009}\\
  \end{tabular}
  \caption{Method comparison}
  \label{tab:results_shapes_b}
  \end{subtable}
   \caption{
   Table (a) shows how quickly the algorithm evaluation converges over 
   inference iterations with the Shapes dataset. Table (b) compares
   segmentation quality to previous work on the Shapes dataset. The AMI score
   is defined in the range from 0 (guessing) to 1 (perfect match).
   The results with a star~(*) are using LayerNorm~\cite{Ba2016} instead of BatchNorm. 
  }
  \label{tab:results_shapes}
\end{table}

\autoref{tab:results_shapes} shows the median performance of Tagger on the Shapes dataset over 20 seeds. 
Tagger is able to achieve very fast convergences, as shown in \autoref{tab:results_shapes_a}. 
Through iterations, the network improves its denoising performances by grouping different objects into different groups. 
Comparing to \citet{Greff2015}, Tagger performs significantly better in terms of AMI score (see \autoref{tab:results_shapes_b}).

\begin{figure}
    \centering
    \includegraphics[width=\textwidth]{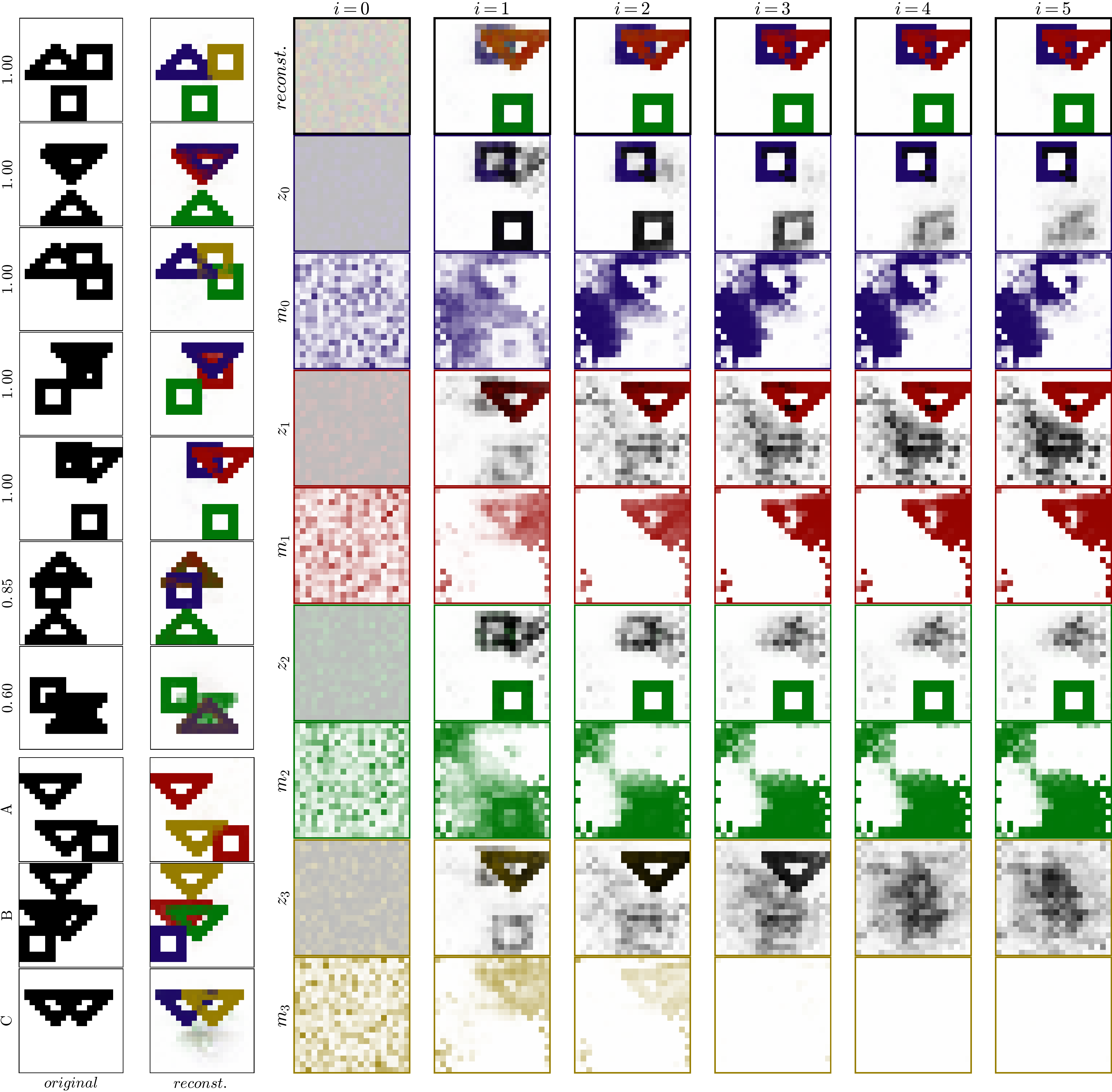}
    \caption{Results for Shapes dataset.
    Left column: 7 examples from the test set along with their resulting groupings in descending AMI score order and
    3 hand-picked examples (A, B, and C) to demonstrate generalization.
    A: Testing 2-group model on 3 object data. B: Testing a 4-group model trained with 3-object data on 4 objects.
    C: Testing 4-group model trained with 3-object data on 2 objects.
    Right column: Illustration of the inference process over iterations for four color-coded groups; $\bf{m}_g$ and $\bf{z}_g$.}
    \label{fig:shapes_qualitative}
\end{figure}

\begin{figure}
	\includegraphics[width=\textwidth]{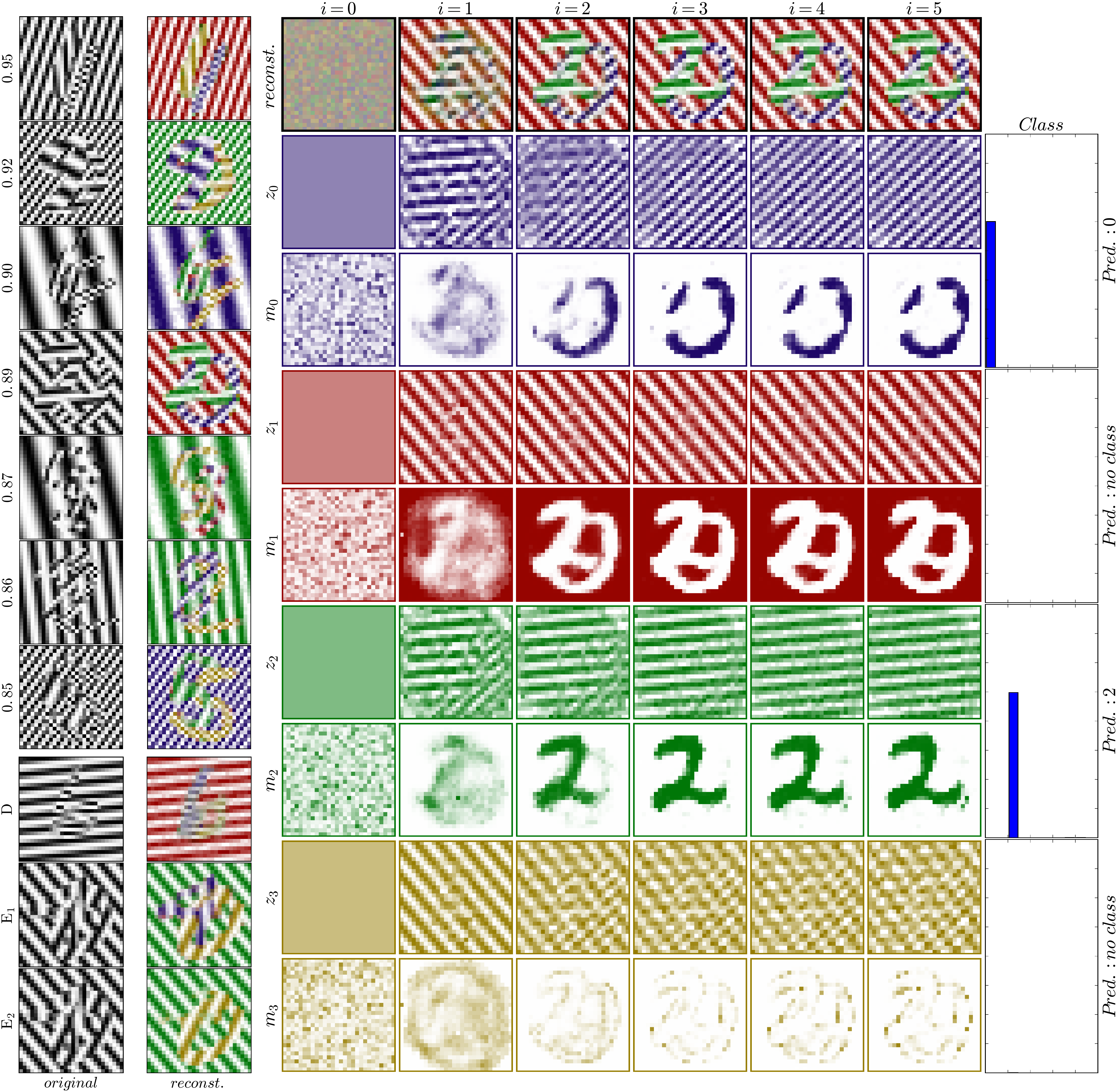}
    \caption{
	Results for the TextureMNIST2 dataset.
    Left column: 7 examples from the test set along with their resulting groupings in descending AMI score order and 3 hand-picked examples (D, E1, E2).
    D: An example from the TextureMNIST1 dataset.
    E1-2: A hand-picked example from TextureMNIST2. E1 demonstrates typical inference, and E2 demonstrates how the system is able to estimate the input when a certain group (topmost digit 4) is removed.
    Right column: Illustration of the inference process over iterations for four color-coded groups; $\bf{m}_g$ and $\bf{z}_g$.
    }
	\label{fig:mnist_qualitative}
\end{figure}

\autoref{fig:shapes_qualitative} and \autoref{fig:mnist_qualitative} qualitatively show the learned unsupervised groupings for the Shapes and textured MNIST datasets.
Tagger uses its TAG mechanism slightly differently for the two datasets.
For Shapes, $\bm{z}_g$ represents filled-in objects and masks $\bm{m}_g$ show which part of the object is actually visible.
For textured MNIST, $\bm{z}_g$ represents the textures and masks $\bm{m}_g$ texture segments.
In the case of the same digit or two identical shapes, Tagger can segment them into separate groups, and hence, it performs instance segmentation.
We used 4 groups for training even though there are only 3 objects in the Shapes dataset and 3 segments in the TexturedMNIST2 dataset. 
The excess group is left empty by the trained system but its presence seems to speed up the learning process.


The hand-picked examples \texttt{A}-\texttt{C} in \autoref{fig:shapes_qualitative} illustrate the robustness of the system when the number of objects changes in the evaluation dataset or when evaluation is performed using fewer groups.

Example $E$ is particularly interesting; $E_1$ shows how the normal evaluation looks like 
but $E_2$ demonstrates how we can remove the topmost digit from the scene and let the
system fill in digit below and the background. 
We do this by setting the corresponding group assignment probabilities $\bm{m}_g$ to a large negative number just before the final softmax over groups in the last iteration.

To solve the textured two-digit MNIST task, the system has to combine texture cues with high-level shape information.
The system first infers the background texture and mask which are finalized on the first iteration.
Then the second iteration typically fixes the texture used for topmost digit, while subsequent iterations clarify the occluded digit and its texture. 
This demonstrates the need for iterative inference of the grouping.

\subsection{Classification}
We investigate the role of grouping for the task of classification.
We evaluate the Tagger against four baseline models on the textured MNIST task.
As our first baseline we use a fully connected network (\emph{FC}) with ReLU activations and batch normalization after each layer.
Our second baseline is a ConvNet (\emph{Conv}) based on Model C from \citet{Springenberg2014}, which has close to state-of-the-art results on CIFAR-10.
We removed dropout, added batch normalization after each layer and replaced the final pooling by a fully connected layer to improve its performance for the task.
Furthermore, we compare with a fully connected Ladder~\cite{Rasmus2015} (FC Ladder) network.

All models use a softmax output and are trained with 50,000 samples to minimize the categorical cross entropy error.
In case there are two different digits in the image (most examples in the TextureMNIST2 dataset), the target is $p=0.5$ for both classes. 
We evaluate the models based on classification errors. For the two-digit case, we score the network based on the two highest predicted classes (top 2).

For Tagger, we first train the system in an unsupervised phase for 150 epochs and then add two fresh randomly initialized layers on top and continue training the entire system end to end using the sum of unsupervised and supervised cost terms for 50 epochs.
Furthermore, the topmost layer has a per-group softmax activation that includes an added 'no class' neuron for groups that do not contain any digit.
The final classification is then performed by summing the softmax output over all groups for the true 10 classes and renormalizing this sum to add up to one.

The final results are summarized in \autoref{tab:results}.
As shown in this table, Tagger performs significantly better than all the fully connected baseline models on both variants, but the improvement is more pronounced for the two-digit case.
This result is expected because for cases with multi-object overlap, grouping becomes more important.
It, moreover, confirms the hypothesis that grouping can help classification and is particularly beneficial for complex inputs.
Remarkably, Tagger, despite being fully connected, is on par with the convolutional baseline for the TexturedMNIST1 dataset and even outperforms it in the two-digit case.
We hypothesize that one reason for this result is that grouping allows for the construction of efficient invariant features already in the low layers without losing information about the assignment of features to objects. 
Convolutional networks solve this problem to some degree by grouping features locally through the use of receptive fields,
but that strategy is expensive and can break down in cases of heavy overlap.

\subsection{Semi-Supervised Learning}

Training TAG does not rely on labels and is therefore directly usable in a semi-supervised context.
For semi-supervised learning, the Ladder~\cite{Rasmus2015} is arguably one of the strongest baselines with SOTA results on 1,000 MNIST and 60,000 permutation invariant MNIST classification.
We follow the common practice of using 1,000 labeled samples and 49,000 unlabeled samples for training Tagger and the Ladder baselines.
For completeness, we also report results of the convolutional (\emph{ConvNet}) and fully-connected (\emph{FC}) baselines trained fully supervised on only 1,000 samples.

From the results in \autoref{tab:results}, it is obvious that all the fully supervised methods fail on this task with 1,000 labels. 
The best result of approximately $52~\%$ error for the single-digit case is achieved by ConvNet, which still performs only at chance level for two-digit classification.
The best baseline result is achieved by the \emph{FC Ladder}, which reaches $30.5~\%$ error for one digit but $68.5~\%$ for TextureMNIST2.

For both datasets, Tagger achieves by far the lowest error rates: $10.5~\%$ and $24.9~\%$, respectively.
Again, this difference is amplified for the two-digit case, where the Tagger with 1,000 labels even outperforms the Ladder baseline with all 50k labels.
This result matches our intuition that grouping can often segment out objects even of an unknown class and thus help select the relevant features for learning.
This is particularly important in semi-supervised learning where the inability to self-classify unlabeled samples can easily mean that the network fails to learn from them at all.

To put these results in context, we performed informal tests with five human subjects. 
The task turned out to be quite difficult and the subjects needed to have regular breaks to be able to maintain focus. 
The subjects improved significantly over training for a few days but there were also significant individual differences. 
The best performing subjects scored around 10~\% error for TextureMNIST1 and 30~\% error for TextureMNIST2. 
For the latter task, the test subject took over 30 seconds per sample.

\begin{table}
{\renewcommand{\arraystretch}{1.2}%
\begin{tabular}{ r l r r l l }
       Dataset & Method & Error 50k & Error 1k & Model details\\
\hline
\hline
TextureMNIST1 & FC MLP        & 31.1 $\pm$ 2.2 &  89.0 $\pm$ 0.2  & 2000-2000-2000 / 1000-1000\\
              & FC Ladder    &  7.2 $\pm$ 0.1 &  30.5 $\pm$ 0.5 & 3000-2000-1000-500-250 &  \\
              & FC Tagger (ours) & \textbf{4.0 $\pm$ 0.3} & \textbf{10.5 $\pm$ 0.9} & 3000-2000-1000-500-250 \\
\hline
              & ConvNet   &  \textbf{3.9 $\pm$ 0.3} &  52.4 $\pm$ 5.3 & based on Model C \cite{Springenberg2014}\\
\hline
\hline
    TextureMNIST2 
   			  & FC MLP     &  55.2 $\pm$ 1.0 &   79.4 $\pm$ 0.3  &  2000-2000-2000 / 1000-1000 \\
              & FC Ladder    &  41.1 $\pm$ 0.2 &  68.5 $\pm$ 0.2 & 3000-2000-1000-500-250 &  \\
              & FC Tagger (ours) & \textbf{7.9 $\pm$ 0.3} & \textbf{24.9 $\pm$ 1.8} & 3000-2000-1000-500-250  \\
\hline
              & ConvNet   &  12.6 $\pm$ 0.4 &   79.1 $\pm$ 0.8 & based on Model C \cite{Springenberg2014} \\

\end{tabular}}
\caption{
Test-set classification errors for textured one-digit MNIST (chance level: 90~\%) and top-2 error on the textured two-digit MNIST dataset (chance level: 80~\%).
We report mean and sample standard deviation over 5 runs. FC = Fully Connected
}
\label{tab:results}
\end{table}


\section{Related work}

Attention models have recently become very popular, and similar to perceptual grouping they help in dealing with complex structured inputs.
These approaches are not, however, mutually exclusive and can benefit from each other.
Overt attention models \cite{Schmidhuber1991,Eslami2016} control a window (fovea) to focus on relevant parts of the inputs.
Two of their limitations are that they are mostly tailored to the visual domain and are usually only suited to objects that are roughly the same shape as the window. 
But their ability to limit the field of view can help to reduce the complexity of the target problem and thus also help segmentation.
Soft attention mechanisms \cite{Schmidhuber1993,Deco2001,Yli-Krekola2009} on the other hand use some form of top-down feedback to suppress inputs that are irrelevant for a given task. 
These mechanisms have recently gained popularity, first in machine translation \cite{Bahdanau2014} and then for many other problems such as image caption generation~\cite{Xu2015}.
Because they re-weigh all the inputs based on their relevance, they could benefit from a perceptual grouping process that can refine the precise boundaries of attention.

Our work is primarily built upon a line of research based on the concept that the brain uses synchronization of neuronal firing to bind object representations together. 
This view was introduced by \citet{vonderMalsburg1981} and has inspired many early works on oscillations in neural networks (see the survey~\cite{vonderMalsburg1995} for a summary). 
Simulating the oscillations explicitly is costly and does not mesh well with modern neural network architectures (but see \cite{Meier2014}).
Rather, complex values have been used to model oscillating activations using the phase as soft tags for synchronization~\cite{Rao2008,Reichert2013}. 	
In our model, we further abstract them by using discretized synchronization slots (our groups).
It is most similar to the models of \citet{Wersing2001}, \citet{Hyvarinen2006} and \citet{Greff2015}.
However, our work is the first to combine this with denoising autoencoders in an end-to-end trainable fashion.

Another closely related line of research~\cite{Saund1995,Ross2006} has focused on multi-causal modeling of the inputs. 
Many of the works in that area \cite{LeRoux2011,Tang2012,Sohn2013,Huang2015} build upon Restricted Boltzmann Machines. 
Each input is modeled  as a mixture model with a separate latent variable for each object.
Because exact inference is intractable, these models approximate the posterior with some form of expectation maximization~\cite{Dempster1977} or sampling procedure.
Our assumptions are very similar to these approaches, but we allow the model to learn the amortized inference directly (more in line with \citet{Goodfellow2013}).

Since recurrent neural networks (RNNs) are general purpose computers, they can in principle implement arbitrary computable types of temporary variable binding~\cite{Schmidhuber1992a, Schmidhuber1993}, unsupervised segmentation~\cite{Schmidhuber1992}, and internal~\cite{Schmidhuber1993} and external attention~\cite{Schmidhuber1991}. 
For example, an RNN with fast weights~\cite{Schmidhuber1993} can rapidly associate or bind the patterns to which the RNN currently attends.
Similar approaches even allow for metalearning~\cite{Schmidhuber1993a}, that is, learning a learning algorithm. 
\citet{Hochreiter2001}, for example, learned fast online learning algorithms for the class of all quadratic functions of two variables.
Unsupervised segmentation could therefore in principle be learned by any RNN as a by-product of data compression or any other given task.

The recurrent architecture most similar to the Tagger is the \emph{Neural Abstraction Pyramid}~(NAP; \citep{Behnke1999}) -- a convolutional neural network augmented with lateral connections which help resolve local ambiguities and feedback connections that allow incorporation of high-level information.
In early pioneering work the NAP was trained for iterative image binarization~\cite{Behnke2003} and iterative image denoising~\cite{Behnke2001}, much akin to the setup we use.
Being recurrent, the NAP layers too, could in principle learn a perceptual grouping as a byproduct.
That does not, however, imply that every RNN will, through learning, easily discover and implement this tool.
The main improvement that our framework adds is an explicit mechanism for the network to split the input into multiple representations and thus quickly and efficiently learn a grouping mechanism.
We believe this special case of computation to be important enough for many real-world tasks to justify this added complexity.


\section{Future Work}

So far we've assumed the groups to represent independent objects or events.
However, this assumption is unrealistic in many cases.
Assuming only \emph{conditional} independence would be considerably more reasonable, and could be implemented by allowing all groups to share the same top-layer of their Ladder network.

The TAG framework assumes just one level of (global) groups, which does not reflect the hierarchical structure of the world.
Therefore, another important future extension is to rather use a hierarchy of local groupings, by using our model as a component of a bigger system.
This could be achieved by collapsing the groups of a Tagger network by summing them together at some hidden layer.
That way this abstract representation could serve as input for another tagger with new groupings at this higher level. 
We hypothesize that a hierarchical Tagger could also represent relations between objects, because they are simply the couplings that remain from the assumption of independent objects.

Movement is a strong segmentation cue and a simple temporal extensions of the TAG framework could be to
allow information to flow forward in time between higher layers, not just via the inputs. 
Iteration would then occur in time alongside the changing inputs.
We believe that these extensions will make it possible to scale the approach to video.


\section{Conclusion}
In this paper, we have argued that the ability to group input elements and internal representations is a powerful tool that can improve a system's ability to handle complex multi-object inputs.
We have introduced the TAG framework, which enables a network to directly learn the grouping and the corresponding amortized iterative inference in a unsupervised manner.
The resulting iterative inference is very efficient and converges within five iterations.
We have demonstrated the benefits of this mechanism for a heavily cluttered classification task, in which our fully connected Tagger even significantly outperformed a state-of-the-art convolutional network.
More impressively, we have shown that our mechanism can greatly improve semi-supervised learning, exceeding conventional Ladder networks by a large margin.
Our method makes minimal assumptions about the data and can be applied to any modality. 
With TAG, we have barely scratched the surface of a comprehensive integrated grouping mechanism, but we already see significant advantages.
We believe grouping to be crucial to human perception and are convinced that it 
will help to scale neural networks to even more complex tasks in the future.

\subsubsection*{Acknowledgments}

The authors wish to acknowledge useful discussions with Theofanis Karaletsos, Jaakko Särelä, Tapani Raiko, and Søren Kaae Sønderby.  
And further acknowledge Rinu Boney, Timo Haanpää and the rest of the Curious AI Company team for their support, computational infrastructure, and human testing. 
This research was supported by the EU project ``INPUT'' (H2020-ICT-2015 grant no. 687795).


\bibliographystyle{icml2016}
{\small
\setlength{\bibsep}{0.3ex}
\bibliography{Tagger}
}

\newpage
\appendix


\section{Supplementary Material}

\subsection{Notation}
\begin{tabular}{ccl}
\toprule
Symbol & Space & Description \\
\midrule
$N$  & $\mathbb{N}$ & input dimensionality \\
$K$  & $\mathbb{N}$ & total number of groups \\
$H$  & $\mathbb{N}$ & input and output dimension of the parametric mapping \\
$i$  & $\mathbb{N}$ & iteration index \\
$j$  & $\{1, \dots, N\}$ & input element index \\
$k$  & $\{1, \dots, K\}$ & group index \\
\midrule
$\bm{x}$         & $\mathbb{R}^{N}$ & input vector with elements $x_j$ \\
$\bm{\tilde{x}}$ & $\mathbb{R}^{N}$ & corrupted input \\
$\bm{z}_k  $ & $\mathbb{R}^{N}$ & the predicted mean of input for group $k$ \\
$\bm{m}_k  $ & $\mathbb{R}^{N}$ & probabilities for each input to be assigned to group $k$ \\
$\bm{\delta z}_k$ & $\mathbb{R}^{N}$ & modeling error for group $k$\\
$L(\bm{m}_k)$     & $\mathbb{R}^{N}$ & group assignment likelihood ratio\\
$C(\bm{x})$ & $\mathbb{R}$ & the training loss for input $\bm{x}$\\
\midrule
$v$ & $\mathbb{R}$ & variance of the input estimate. Only used in the continuous case \\
$W_h$ & $\mathbb{R}^{H \times 4N}$ & Projection weights from tagger inputs to ladder inputs $\bm{h}$ \\
$W_u$ & $\mathbb{R}^{2N \times H}$ & Projection weights from ladder output to $\bm{z}$ and $\bm{m}$ \\
$\Theta$ & & Contains all parameters of the ladder\\
\midrule
$f()$ & & rectified linear activation function \\
$g()$ & & logistic sigmoid activation function \\
$\text{softmax}()$ & & elementwise softmax over the groups   \\
\midrule
$G_j$ & & Latent random variable that encodes which group $x_j$ belongs to.\\
$g_{k,j}$ & & Shorthand for $G_j = k$. Mostly used for $p(g_{k,j}) = p(G_j = k)$.\\
$\bm{g}$ & & a vector of all $g_j$. \\
$p(\bm{x} \mid \bm{\tilde{x}})$ & & posterior of the data given the corrupted data \\
$q(\bm{x})$ &  & learnt approximation of $p(\bm{x} \mid \bm{\tilde{x}})$ \\
$q(x_j \mid g_{k,j})$ &  &  Shorthand for $q(x_j \mid G_j = k)$ \\
\bottomrule
\end{tabular}

\begin{figure}
\centering
\includegraphics[width=1.0\textwidth]{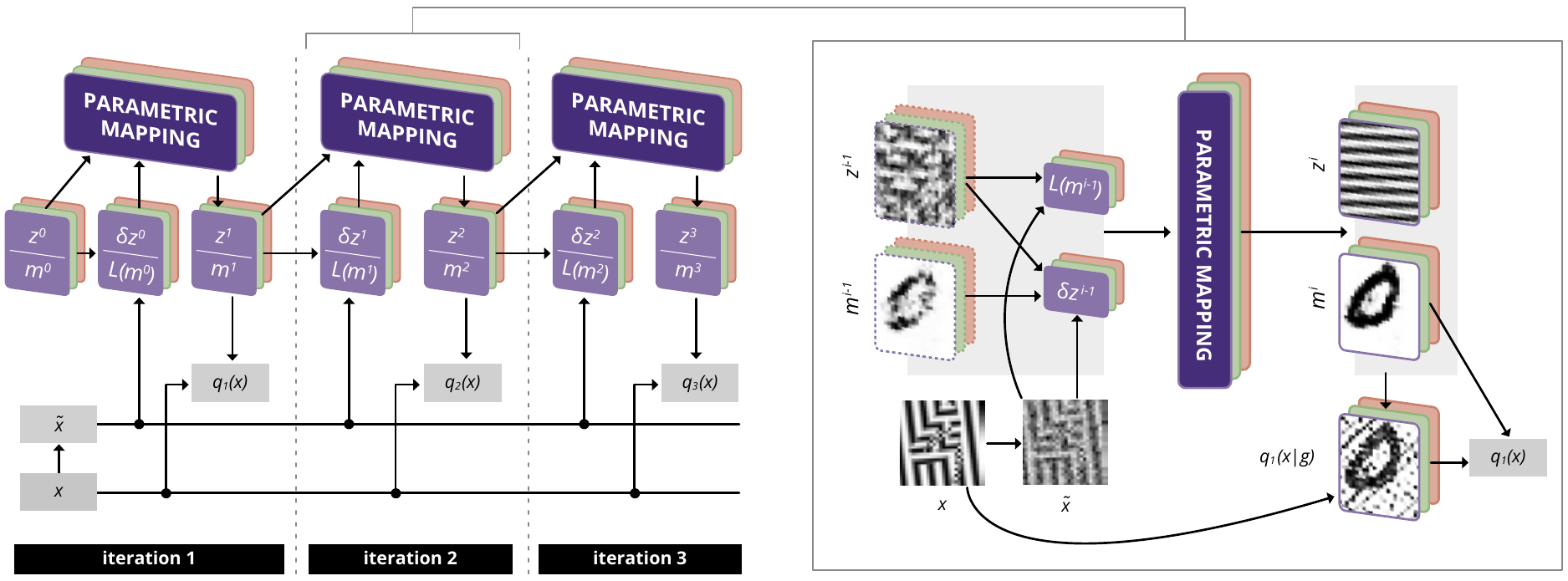}
\caption{Illustration of the TAG framework used for training. Left: The system
learns by denoising its input over iterations using several groups to distribute the
representation.
Each group, represented by several panels of the same color, maintains its own estimate of reconstructions $\bm{z}^i$ of the input, and corresponding masks $\bm{m}^i$, which encode the parts of the input that this group is responsible for representing. 
These estimates are updated over iterations by the same network, that is, each group and iteration share the weights of the network and only the inputs to the network differ.
In the case of images, $\bm{z}$ contains pixel-values.
Right: In each iteration $\bm{z}^{i-1}$ and $\bm{m}^{i-1}$ from the previous iteration, 
are used to compute a likelihood term $L(\bm{m}^{i-1})$ and modeling error $\delta \bm{z}^{i-1}$. 
These four quantities are fed to the parametric mapping to produce $\bm{z}^i$ and $\bm{m}^i$ for the next iteration. 
During learning, all inputs to the network are derived from the corrupted input as shown here. 
The unsupervised task for the network is to learn to denoise, i.e. output an estimate $q(\bm{x})$ of the original clean input.}
\end{figure}

\subsection{Input}
\label{sec:inputnoise}
In its basic form (without supervision) Tagger receives as input only a datapoint $\bm{x}$.
It corresponds to either a binary vector or a real-valued vector and is then corrupted with either bitflip or Gaussian noise.
The training objective is the removal of this noise. 

\paragraph{Bitflip Noise}
In the case of binary inputs we use bitflip noise for corruption:
\[
\bm{\tilde{x}} = \bm{x} \oplus \mathcal{B}(\beta),
\]
where $\oplus$ denotes componentwise XOR, and $\mathcal{B}(\beta)$ is Bernoulli distributed noise with probability $\beta$. In our experiments on the Shapes dataset we use $\beta = 0.2$.

\paragraph{Gaussian Noise}
If the inputs are real-valued, we corrupt it using Gaussian noise:
\[
\bm{\tilde{x}} = \bm{x} + \mathcal{N}(0, \sigma^2),
\]
where $\sigma$ is the standard deviation of the input noise. We used $\sigma_{input} = 0.2$.

\newpage
\subsection{Group Assignments}
\label{sec:m}
Within the TAG framework the group assignment is represented by the $K$ vectors $\bm{m}_k$ which contain one entry for each input element or pixel.
These entries $m_{k,j} = q(g_{k,j})$ of $\bm{m}_k$ represent the discreet probability distribution over $K$ groups for each input $x_j$.
They therefore sum up to one:
\begin{equation}
  \sum_{k=1}^{K} m_{k, j} = 1\quad \text{ for all } j=1\dots N
\label{eq:m_sum}
\end{equation}

\paragraph{Initialization}
Similar to expectation maximization, the group assignment is initialized randomly, but such that \autoref{eq:m_sum} holds.
So we first sample an auxiliary $m'_{k,j}$ from a standard Gaussian distribution and then normalize it using a softmax:
\begin{align}
m'_{k,j} &\sim \mathcal{N}(0, 1)\\
m_{k,j}  &= \frac{e^{m'_{k,j}}}{\sum_{h=1}^K e^{m'_{h, j}}}\\
\end{align}

\subsection{Predicted Inputs}
\label{sec:z}
Tagger maintains an input reconstruction $\bm{z}_k$ for each group $k$.

\paragraph{Binary Case}
In the binary case we use a sigmoid activation function on $\bm{z}_k$ and interpret it directly as the probability 
\begin{equation}
  \operatorname{sigmoid}(\bm{z}_k) = q(\bm{x} = \bm{1} | \bm{g}_{k}).
\end{equation}
We can use it to compute $\bm{\tilde{z}_k} = q(\bm{\tilde{x}}| \bm{g}_k)$ which will be used for the modeling error (\autoref{sec:deltaz}) and the group likelihood:
\begin{align}
  q(\bm{\tilde{x}} = 1 | \bm{g}_k) &= \sum_{\bm{x}} q(\bm{\tilde{x}} | \bm{x}, \bm{g}_k) q(\bm{x} | \bm{g}_k) \\
                &= \sum_{\bm{x}} q(\bm{\tilde{x}} | \bm{x}) q(\bm{x} | \bm{g}_k) \\
                &= \sum_{\bm{x}} \left( \bm{x} (1-\beta) + (1-\bm{x}) \beta \right) q(\bm{x} | \bm{g}_k) \\
                &= \sum_{\bm{x}} \left( \bm{x} (1-2\beta) + \beta\right) \bm{z}_k \\
                &= \beta (1 - \bm{z}_k) + (1-\beta) \bm{z}_k \\
                &= \bm{z}_k (1 - 2 \beta) + \beta
\end{align}
Therefore we have:
\begin{equation}
	\bm{\tilde{z}}_k = \bm{\tilde{x}} (\bm{z}_k (1 - 2 \beta) + \beta) + (1 - \bm{\tilde{x}}) (1 - \bm{z}_k (1 - 2 \beta) - \beta)
\end{equation}

\paragraph{Continuous Case}
For the continuous case we interpret $\bm{z}_k$ as the means of an isotropic Gaussian with learned variance~$v$: 
\begin{equation}
    q(\bm{x} | \bm{g}_k) = \mathcal{N}(\bm{x}; \bm{z}_k, v\bm{I}) = \frac{1}{\sqrt{2\pi v}} e^{\frac{(\bm{x} - \bm{z}_k)^2}{2v}}
\end{equation}
Using the additivity of Gaussian distributions we directly get:
\begin{equation}
    \bm{\tilde{z}}_k = q(\bm{\tilde{x}} | \bm{g}_k) = \mathcal{N}(\bm{\tilde{x}}; \bm{z}_k, (v + \sigma^2)\bm{I})
\end{equation}

\paragraph{Initialization}
For simplicity we initialize all $\bm{z}_k$ to the expectation of the data for all $k$.
In our experiments these values are $0.5$ for the TextureMNIST datasets and $0.26$ for the Shapes dataset.

\subsection{Modeling Error}
\label{sec:deltaz}

As explained in \autoref{sec:synthesis}, $\bm{\delta z}$ carries information about the remaining modeling error.
During training as a denoiser, we can only allow information about the corrupted $\bm{\tilde{x}}$ as inputs but not about the original clean $\bm{x}$. 
Therefore, we use the derivative of the cost on the corrupted input as helpful information for the parametric mapping. Since we work with the input elements individually we skip the index $j$ in the following:
\begin{equation}
    {\delta z}_k \propto -\partial C({\tilde{x}})/\partial {z}_k.
\end{equation}
More precisely for a single iteration (omitting the index $i$) we have::
\begin{align}
  \delta {z}_k &= -\frac{\partial {C({\tilde{x}})}}{\partial  {z}_{k}}\\ 
      &= \frac{\partial}{\partial  z_{k}} \log \left( \sum_h q(\tilde{x} | g_{h}) q(g_{h}) \right)  \\ 
      &= \frac{1}{\sum_h q(\tilde{x} | z_{h}) q(g_{h})}  \frac{\partial \sum_h q(\tilde{x} | z_{h}) q(g_{h})}{\partial  z_{k}} \\
      &= \frac{1}{\sum_h \tilde{z}_h m_{h}} \frac{\partial \tilde{z}_k }{\partial  z_{k}} m_{k}   \\ 
\end{align}

\paragraph{Continuous Case}
For the continuous case this gives us:
\begin{align}
  \delta {z}_k &= \frac{1}{\sum_h \tilde{z}_h m_{h}} \frac{\partial \tilde{z}_k }{\partial  z_{k}} m_{k} \\ 
      &= \frac{1}{\sum_h \tilde{z}_h m_{h}} \frac{\tilde{x} - z_k}{\sigma^2 + v} \tilde{z}_k m_{k}   \\     
      &\propto (\tilde{x} - z_k) m_k \tilde{z}_k
\end{align}
Note that since the network will multiply its inputs with weights, we can always omit any constant multipliers.
\newpage
\paragraph{Binary Case}
Let us denote the corruption bit-flip probability by $\beta$ and define
\[
\xi_k := q(\tilde{x}=1 | g_k) = (1-2\beta)z_g+\beta \, .
\]
Then we get:
\[
\tilde{z}_k = \tilde{x} \xi_k + (1 - \tilde{x}) (1 - \xi_k) 
\]
and thus:
\begin{align}
  \delta {z}_k &= \frac{1}{\sum_h \tilde{z}_h m_{h}} \frac{\partial \tilde{z}_k }{\partial  z_{k}} m_{k} \\ 
  &=\frac{(\tilde{x}(1-2\beta)-(1-\tilde{x})(1-2\beta))m_{k}}{\sum_{h}(\tilde{x}\xi_{h}+(1-\tilde{x})(1-\xi_{h}))m_{h}}
\end{align}
which simplifies for $\tilde{x}=1$ as
\[
=\frac{(1-2\beta)m_{k}}{\sum_{h}\xi_{h} m_{h}}\approx-\frac{m_{k}}{\sum_{h}\xi_{h} m_{h}}
\]
and for $\tilde{x}=0$ as
\[
=\frac{(1-2\beta)m_{k}}{1-\sum_{h}\xi_{h} m_{h}} \approx \frac{m_{k}}{1-\sum_{h}\xi_{h} m_{h}}=\frac{m_{k}}{\sum_{h}\xi_{h}m_{h}-1}
\]
Putting it back together:
\[
\delta {z}_k =\frac{m_{k}}{\sum_{h}\xi_{h} m_{h}-1 + \tilde{x}}
\]

\subsection{Ladder Modifications}
We mostly used the specifications of the Ladder network as described by \citet{Rasmus2015}, but there are some minor modifications we made to fit it to the TAG framework. 
We found that the model becomes more stable during iterations when we added a sigmoid function to the gating variable $v$~\citep[Equation 2]{Rasmus2015} used in all the decoder layers with continuous outputs.
None of the noise sources or denoising costs were in use (i.e., $\lambda_l=0$ for all $l$ in Eq.~3 of Ref.~\cite{Rasmus2015}), but Ladder's classification cost ($C_c$ in Ref.~\cite{Rasmus2015}) was added to the Tagger's cost for the semi-supervised tasks.

All four inputs ($\bm{z}^i_k$, $\bm{m}^i_k$, $\delta \bm{z}^{i}_k$, and $L(\bm{m}^{i}_k)$) were concatenated and projected to a hidden representation that served as the input layer of the Ladder Network. 
Subsequently, the values for the next iteration were simply read from the reconstruction ($\hat{\bm{x}}$ in Ref.~\cite{Rasmus2015}) and projected linearly into $\bm{z}^{i+1}_k$ and via softmax to $\bm{m}^{i+1}_k$ to enforce the conditions in \autoref{eq:m_sum}. 
For the binary case, we used a logistic sigmoid activation for $\bm{z}^{i+1}_k$.

\newpage
\subsection{Pseudocode} 
\label{sec:pseudocode}
In this section we put it all together and provide the pseudocode for running Tagger both on binary (\autoref{alg:tagger_bin}) and real-valued inputs (\autoref{alg:tagger_cont}).
The provided code shows the steps needed to run for $T$ iterations on a single example $\bm{x}$ using $G$ groups.
Here we use three activation functions: $f(x) = \max(x, 0)$ is the rectified linear function, 
$g(x)=\frac{1}{1+e^{-x}}$ is the logistic sigmoid, and $\text{softmax}(x)_g = \frac{e^{x_g}}{\sum_{h=1}^{G} e^{x_h}}$ is a softmax operation over the groups.
All three include a batch-normalization operation, which we omitted for clarity.
Only the forward pass for a single example is shown, but derivatives of the cost $C$ wrt. parameters $v$, $W_h$, $W_u$ and $\Theta$ are computed using regular backpropagation through time.
For training we use ADAM with a batch-size of $100$.

\begin{algorithm}[H]
  \KwData{$\bm{x}, K, T, \sigma, v, W_h, W_u\Theta$}
  \KwResult{$\bm{z}^T, \bm{m}^T,  C$}
  \Begin(Initialization:){
    $\bm{\tilde{x}} \leftarrow \bm{x} + \mathcal{N}(\bm{0}, \sigma^2\bm{I}) $\;
    $\bm{m}^0 \leftarrow \text{softmax}(\mathcal{N}(\bm{0}, \bm{I}))$\;
    $\bm{z}^0 \leftarrow E[\bm{x}]$\;
  }
  \For{$i = 0 \dots T-1$}{
    \For{$k = 1 \dots K$}{
      $\bm{\tilde{z}}_k \leftarrow \mathcal{N}(\bm{\tilde{x}}; \bm{z}^i_k, (v + \sigma^2) \bm{I})$\;
      $\bm{\delta z^i_k} \leftarrow (\bm{\tilde{x}} - \bm{z}_k^i) \bm{m}^i_k \bm{\tilde{z}}_k $\;
      $L(\bm{m}^i_k) \leftarrow \frac{\bm{\tilde{z}}_k}{\sum_h \bm{\tilde{z}}_h}$ \;
      $\bm{h}_k \leftarrow f(W_h \left[ \bm{z}^i_k, \bm{m}^i_k, \bm{\delta z}^i_k, L(\bm{m}^i_k) \right])$\;
      $[ \bm{z}^{i+1}_k, \bm{m}^{i+1}_k ] \leftarrow W_u \text{Ladder}(\bm{h}_k, \Theta)$\; 
    }
    $\bm{m}^{i+1} \leftarrow \operatorname{softmax} (\bm{m}^{i+1})$\;
    $q_{i+1}(\bm{x}) \leftarrow \sum_{k=1}^K \mathcal{N}(\bm{x}; \bm{z}^{i+1}_k, v\bm{I}) \bm{m}^{i+1}$\;
  }
 $C \leftarrow - \sum_{i=1}^T \log q_i(\bm{x}) $\;
 \vspace{0.5em}
 \caption{Pseudocode for running Tagger on a single real-valued example $\bm{x}$.}
 \label{alg:tagger_cont}
\end{algorithm}

\begin{algorithm}[H]
  \KwData{$\bm{x}, K, T, \beta, W_h, W_u, \Theta$}
  \KwResult{$\bm{z}^T, \bm{m}^T,  C$}
  \Begin(Initialization:){
    $\bm{\tilde{x}} \leftarrow \bm{x} \oplus \mathcal{B}(\beta) $\;
    $\bm{m}^0 \leftarrow \text{softmax}(\mathcal{N}(\bm{0}, \bm{I}))$\;
    $\bm{z}^0 \leftarrow E[\bm{x}]$\;
  }
  \For{$i = 0 \dots T-1$}{
    \For{$k = 1 \dots K$}{
      $\bm{\xi_k} \leftarrow \bm{z}^i(1-2\beta) + \beta$\;
      $\bm{\delta z^i_k} \leftarrow \frac{\bm{m}^i_k}{\sum_h \bm{\xi}_h \bm{m}^i_h - \bm{1} + \bm{\tilde{x}}} $\;
    $L(\bm{m^i}) \leftarrow \frac{\bm{\tilde{x}} \bm{\xi}_k + (1 - \bm{\tilde{x}})(1 - \bm{\xi}_k)}{\sum_h \bm{\tilde{x}} \bm{\xi}_h + (1 - \bm{\tilde{x}})(1 - \bm{\xi}_h)}$ \;
      
      $\bm{h}_k \leftarrow f(W_h \left[ \bm{z}^i_k, \bm{m}^i_k, \bm{\delta z}^i_k, L(\bm{m}^i_k) \right])$\;
      $[ \bm{z}^{i+1}_k, \bm{m}^{i+1}_k] \leftarrow W_u \text{Ladder}(\bm{h}_k, \Theta)$\; 
    }
    $\bm{m}^{i+1} \leftarrow \operatorname{softmax} (\bm{m}^{i+1})$\;
    $q_{i+1}(\bm{x}) \leftarrow \sum_{k=1}^K \mathcal{N}(\bm{x}; \bm{z}^{i+1}_k, v\bm{I}) \bm{m}^{i+1}$\;
  }
 $C \leftarrow - \sum_{i=1}^T \log q_i(\bm{x}) $\;
 \vspace{0.5em}
 \caption{Pseudocode for running Tagger on a single binary example $\bm{x}$.}
 \label{alg:tagger_bin}
\end{algorithm}

\end{document}